\documentclass[a4paper,10pt]{article}
\usepackage{graphicx}
\begin{document}

\title{\bf Accomplishable Tasks in Knowledge Representation}
\author{Keehang Kwon and Mi-Young Park\\
\sl \small Faculty of Computer Engineering, DongA  University\\
\sl \small 840 Hadan Saha, 604-714 Busan, Korea\\
\small  \{ khkwon, openmp \}@dau.ac.kr}
\date{}
\maketitle

%% <local definitions here>

\makeatletter
\long\def\@makemyfntext#1{$^{\rm *}\ $ #1}

\long\def\@myfootnotetext#1{\insert\footins{\footnotesize
    \interlinepenalty\interfootnotelinepenalty 
    \splittopskip\footnotesep
    \splitmaxdepth \dp\strutbox \floatingpenalty \@MM
    \hsize\columnwidth \@parboxrestore
   \edef\@currentlabel{\csname p@footnote\endcsname\@thefnmark}\@makemyfntext
    {\rule{\z@}{\footnotesep}\ignorespaces
      #1\strut}}}

\def\myfootnotetext{\@ifnextchar
     [{\@xfootnotenext}{\xdef\@thefnmark{\thempfn}\@myfootnotetext}}
\makeatother
%\setstretch{2}

\newenvironment{describe}{\begin{list}{}{\setlength\leftmargin{80pt}}\setlength\labelsep{10pt}\setlength\labelwidth{70pt}}{\end{list}}

\newenvironment{flag}{\begin{list}{\makebox[20pt]{\hss$\circ$\enspace}}
                                  {\labelwidth 20pt}}{\end{list}}

%% js \newtheorem{proposition}{Proposition}

%% js \newenvironment{proof}
     %% js {\begin{trivlist}\item[]{\bf Proof. }}%
     %% js {\\* \hspace*{\fill} $\Box$\end{trivlist}}

\newenvironment{numberedlist}
{\begin{list}{\makebox[20pt]{\hss(\arabic{itemno})\enspace}}
             {\usecounter{itemno}\labelwidth 20pt}}{\end{list}}

\newenvironment{alphabetlist}
{\begin{list}{\makebox[20pt]{\hss(\alph{itemno1})\enspace}}
             {\usecounter{itemno1}\labelwidth 20pt}}{\end{list}}

\newenvironment{romanlist}
{\begin{list}{\makebox[20pt]{\hss(\roman{itemno2})\enspace}}
             {\usecounter{itemno2}\labelwidth 20pt}}{\end{list}}

\newcounter{itemno}

\newcounter{itemno1}

\newcounter{itemno2}
\newcounter{lemma}
\newcounter{exno}

\newcounter{defno}

%\newcounter{exno}[section]

%\newcounter{defno}[section]

%\newtheorem{defn}{Definition}[section]

%\newtheorem{ex}[defn]{Example}

%% js \newtheorem{lemma}{Lemma}

%% js \newtheorem{theorem}[lemma]{Theorem}

\newenvironment{defn}{\refstepcounter{defno}\medskip \noindent {\bf
Definition \thedefno.\ }}{\medskip}

\newenvironment{ex}{\refstepcounter{exno}\medskip \noindent {\bf
Example \theexno.\ }}{\medskip}

\newenvironment{millerexample}{
 \begingroup \begin{tabbing} \hspace{2em}\= \hspace{5em}\= \hspace{5em}\=
\hspace{5em}\= \kill}{
 \end{tabbing}\endgroup}

\newenvironment{wideexample}{
 \begingroup \begin{tabbing} \hspace{2em}\= \hspace{10em}\= \hspace{10em}\=
\hspace{10em}\= \kill}{
 \end{tabbing}\endgroup}

\newcommand{\sep}{\;\vert\;}

\newcommand{\ra}{\rightarrow}
\newcommand{\app}{\ }
\newcommand{\appt}{\ }
\newcommand{\tup}[1]{\langle\nobreak#1\nobreak\rangle}

\newcommand{\hu}{{\cal H}^+}
\newcommand{\Free}{{\cal F}}
\newcommand{\oprove}{\vdash\kern-.6em\lower.7ex\hbox{$\scriptstyle O$}\,}
\newcommand{\true}{\top}

\newcommand{\Dscr}{{\cal D}}
\newcommand{\Pscr}{{\cal P}}
\newcommand{\Gscr}{{\cal G}}
\newcommand{\Fscr}{{\cal F}}
\newcommand{\Vscr}{{\cal V}}
\newcommand{\Uscr}{{\cal U}}
\newcommand{\pderivation}{{\cal P}\kern -.1em\hbox{\rm -derivation}}
\newcommand{\pderivationl}{{\cal P}\kern -.1em\hbox{\em -derivation}}
\newcommand{\pderivable}{{\cal P}\kern -.1em\hbox{\rm -derivable}}
\newcommand{\pderivablel}{{\cal P}\kern -.1em\hbox{\em -derivable}}
\newcommand{\pderivations}{{\cal P}\kern -.1em\hbox{\rm -derivations}}
\newcommand{\pderivability}{{\cal P}\kern -.1em\hbox{\rm -derivability}}
\newcommand{\eqm}[1]{=_{\scriptscriptstyle #1}}
\newcommand\subsl{\preceq}
\newcommand{\fnrestr}{\uparrow}

\newcommand{\match}{{\rm MATCH}}
\newcommand{\triv}{{\rm TRIV}}
\newcommand{\imit}{{\rm IMIT}}
\newcommand{\proj}{{\rm PROJ}}
\newcommand{\simpl}{{\rm SIMPL}}
\newcommand{\failed}{{\bf F}}

\newcommand{\Dsiginst}[1]{{[#1]_\Sigma}}
\newcommand{\Psiginst}[1]{{[#1]_\Sigma}}
\newcommand{\lnorm}{{\lambda}norm}
\newcommand{\seq}[2]{#1 \supset #2}
\newcommand{\dseq}[2]{#1_1,\ldots,#1_{#2}}

\newcommand{\all}{\forall}
\newcommand{\some}{\exists}
\newcommand{\lambdax}[1]{\lambda #1\,}
\newcommand{\somex}[1]{\some#1\,}
\newcommand\allx[1]{\all#1\,}

\newcommand{\subs}[3]{[#1/#2]#3}
\newcommand{\rep}[3]{S^{#2}_{#1}{#3}}
\newcommand{\ie}{{\em i.e.}}
\newcommand{\eg}{{\em e.g.}}

% These are the annotations used with inference figures
\newcommand{\lbotr}{$\bot$-R}
\newcommand{\ldbotr}{\bot\mbox{\rm -R}}
\newcommand{\landl}{$\land$-L}
\newcommand{\ldandl}{\land\mbox{\rm -L}}
\newcommand{\landr}{$\land$-R}
\newcommand{\ldandr}{\land\mbox{\rm -R}}
\newcommand{\lorl}{$\lor$-L}
\newcommand{\ldorl}{\lor\mbox{\rm -L}}
\newcommand{\lorr}{$\lor$-R}
\newcommand{\ldorr}{\lor\mbox{\rm -R}}
\newcommand{\limpl}{$\supset$-L}
\newcommand{\ldimpl}{\supset\mbox{\rm -L}}
\newcommand{\limpr}{$\supset$-R}
\newcommand{\ldimpr}{\supset\mbox{\rm -R}}
\newcommand{\lnegl}{$\neg$-L}
\newcommand{\ldnegl}{\neg\mbox{\rm -L}}
\newcommand{\ldnegr}{\neg\mbox{\rm -R}}
\newcommand{\lalll}{$\forall$-L}
\newcommand{\ldalll}{\forall\mbox{\rm -L}}
\newcommand{\lallr}{$\forall$-R}
\newcommand{\ldallr}{\forall\mbox{\rm -R}}
\newcommand{\lsomel}{$\exists$-L}
\newcommand{\ldsomel}{\exists\mbox{\rm -L}}
\newcommand{\lsomer}{$\exists$-R}
\newcommand{\ldsomer}{\exists\mbox{\rm -R}}
\newcommand{\ldlamlr}{\lambda}
\newcommand{\sequent}[2]{\hbox{{$#1\ \longrightarrow\ #2$}}}
\newcommand{\prog}[2]{\hbox{{$#1\ \supset\ #2$}}}
\newcommand{\run}{\Gamma}

\newcommand{\Ibf}{{\bf I}}
\newcommand{\Cbf}{{\bf C}} 
\newcommand{\Cbfpr}{{\bf C'}}

\newcommand{\cprove}{\vdash_C}
\newcommand{\iprove}{\vdash_I}

\newsavebox{\lpartfig}
\newsavebox{\rpartfig}

% From the hohh section

\newenvironment{exmple}{
 \begingroup \begin{tabbing} \hspace{2em}\= \hspace{3em}\= \hspace{3em}\=
\hspace{3em}\= \hspace{3em}\= \hspace{3em}\= \kill}{
 \end{tabbing}\endgroup}
\newenvironment{example2}{
 \begingroup \begin{tabbing} \hspace{8em}\= \hspace{2em}\= \hspace{2em}\=
\hspace{10em}\= \hspace{2em}\= \hspace{2em}\= \hspace{2em}\= \kill}{
 \end{tabbing}\endgroup}

\newenvironment{example}{
\begingroup  \begin{tabbing} \hspace{2em}\= \hspace{3em}\= \hspace{3em}\=
\hspace{3em}\= \hspace{3em}\= \hspace{3em}\= \hspace{3em}\= \hspace{3em}\= 
\hspace{3em}\= \hspace{3em}\= \hspace{3em}\= \hspace{3em}\= \kill}{
 \end{tabbing} \endgroup }

\newcommand{\sand}{sand} % choice disjunction
\newcommand{\pand}{pand} % choice disjunction
\newcommand{\cor}{cor} % choice disjunction

\newcommand{\lb}{\langle}
\newcommand{\rb}{\rangle}
\newcommand{\pr}{prov}
\newcommand{\prG}{intp}
\newcommand{\prSG}{intp_E}
\newcommand{\intp}{intp_o}
\newcommand{\prove}{exec} % choice conjunction
\newcommand{\np}{invalid} % choice conjunction
\newcommand{\Ra}{\supset}  
\newcommand{\Cscr}{{\cal C}}
\newcommand{\seqweb}{SProlog}
\newcommand{\sprog}{{SProlog}}

\newtheorem{theorem}[lemma]{Theorem}

\newtheorem{proposition}[lemma]{Proposition}

\newtheorem{corollary}[lemma]{Corollary}
\newenvironment{proof}
     {\begin{trivlist}\item[]{\it Proof. }}%
     {\\* \hspace*{\fill} \end{trivlist}}

\newcommand{\seqand}{\prec}
\newcommand{\seqor}{\cup}
\newcommand{\seqandq}[2]{\prec_{#1}^{#2}}
\newcommand{\parandq}[2]{\land_{#1}^{#2}}
\newcommand{\exq}[2]{\exists_{#1}^{#2}}
\newcommand{\ext}{intp_G}
\newcommand{\gneg}{\neg} % negation

\newcommand{\mlc}{\wedge} % parallel conjunction

\newcommand{\mld}{\vee} % parallel disjunction

\newcommand{\mli}{\rightarrow} % basic reduction

\newcommand{\mla}{\mbox{{\Large $\wedge$}}} % parallel universal quantifier

\newcommand{\mle}{\mbox{{\Large $\vee$}}} % parallel existential quantifier

\newcommand{\pst}{\mbox{\raisebox{-0.01cm}{\scriptsize $\wedge$}\hspace{-4pt}\raisebox{0.16cm}{\tiny $\mid$}\hspace{2pt}}} % parallel recurrence

\newcommand{\pcost}{\mbox{\raisebox{0.12cm}{\scriptsize $\vee$}\hspace{-4pt}\raisebox{0.02cm}{\tiny $\mid$}\hspace{2pt}}} % parallelcorecurrence

\newcommand{\pintimpl}{\mbox{\hspace{2pt}\raisebox{0.033cm}{\tiny $>$}\hspace{-0.18cm} \raisebox{-0.043cm}{\large --}\hspace{2pt}}} % parallel-recurrence-based reduction

\newcommand{\cla}{\mbox{\large $\forall$}} % blind universal quantifier

\newcommand{\cle}{\mbox{\large $\exists$}} % blind existential quantifier

\newcommand{\adc}{\sqcap} % choice conjunction

\newcommand{\add}{\sqcup} % choice disjunction

\newcommand{\adi}{\sqsupset} % choice implication

\newcommand{\ada}{\mbox{\Large $\sqcap$}} % choice universal quantifier

\newcommand{\ade}{\mbox{\Large $\sqcup$}} % choice existential quantifier

\newcommand{\st}{\mbox{\raisebox{-0.05cm}{$\circ$}\hspace{-0.13cm}\raisebox{0.16cm}{\tiny $\mid$}\hspace{2pt}}} % branching recurrence

\newcommand{\cost}{\mbox{\raisebox{0.12cm}{$\circ$}\hspace{-0.13cm}\raisebox{0.02cm}{\tiny $\mid$}\hspace{2pt}}} 
\newcommand{\intimpl}{\mbox{\hspace{2pt}$\circ$\hspace{-0.14cm} \raisebox{-0.043cm}{\Large --}\hspace{2pt}}} % branching-recurrence-based reduction

\newcommand{\sqc}{\mbox{\small \raisebox{0.0cm}{$\bigtriangleup$}}} % sequential conjunction

\newcommand{\sqd}{\mbox{\small \raisebox{0.049cm}{$\bigtriangledown$}}} % sequential disjunction

\newcommand{\sqa}{\mbox{\large \raisebox{0.0cm}{$\bigtriangleup$}}} % sequential universal quantifier

\newcommand{\sqe}{\mbox{\large \raisebox{0.07cm}{$\bigtriangledown$}}} % sequential existential quantifier

\newcommand{\sst}{\mbox{\raisebox{-0.07cm}{\scriptsize $-$}\hspace{-0.2cm}$\pst$}} % sequential recurrence

\newcommand{\scost}{\mbox{\raisebox{0.20cm}{\scriptsize $-$}\hspace{-0.2cm}$\pcost$}} % sequential corecurrence

\newcommand{\sintimpl}{\mbox{\hspace{2pt}\raisebox{0.033cm}{\tiny $ | \hspace{-4pt} >$}\hspace{-0.14cm} \raisebox{-0.039cm}{\large --}\hspace{2pt}}} % sequential-recurrence-based reduction

\newcommand{\cst}{{\mbox{\raisebox{-0.05cm}{$\circ$}\hspace{-0.13cm}\raisebox{0.16cm}{\tiny $\mid$}\hspace{1pt}}}^{\aleph_0}} % countable recurrence

\newcommand{\ccost}{{\mbox{\raisebox{0.12cm}{$\circ$}\hspace{-0.13cm}\raisebox{0.02cm}{\tiny $\mid$}\hspace{1pt}}}^{\aleph_0}} % countable corecurrence

\newcommand{\cintimpl}{{\mbox{\hspace{2pt}$\circ$\hspace{-0.14cm} \raisebox{-0.043cm}{\Large --}\hspace{1pt}}}^{\aleph_0}} % countable-recurrence-based reduction

\newcommand{\fintimpl}{\mbox{\hspace{2pt}$\bullet$\hspace{-0.14cm} \raisebox{-0.058cm}{\Large --}\hspace{-6pt}\raisebox{0.008cm}{\scriptsize $\wr$}\hspace{-1pt}\raisebox{0.008cm}{\scriptsize $\wr$}\hspace{4pt}}} % dfb-reduction

\newcommand{\sfbr}{\mbox{\hspace{2pt}$\bullet$\hspace{-0.14cm} \raisebox{-0.058cm}{\Large --}\hspace{-6pt}\hspace{1pt}\raisebox{0.008cm}{\scriptsize $\wr$}\hspace{4pt}}} % sfb-reduction

\newcommand{\fbr}{\mbox{\hspace{2pt}$\bullet$\hspace{-0.14cm} \raisebox{-0.058cm}{\Large --}\hspace{2pt}}} % fb-reduction

¡¡

\newcommand{\tgd}{\mbox{\hspace{2pt}$\vee$\hspace{-1.29mm}\raisebox{0.1mm}{\rule{0.13mm}{2mm}}\hspace{5pt}}} % toggling disjunction

\newcommand{\tgc}{\mbox{\hspace{2pt}$\wedge$\hspace{-1.29mm}\raisebox{0.02mm}{\rule{0.13mm}{2mm}}\hspace{5pt}}} % toggling conjunction

\newcommand{\tge}{\hspace{1pt}\mbox{\Large $\vee$\hspace{-1.84mm}\raisebox{0.1mm}{\rule{0.13mm}{3.0mm}}\hspace{6pt}}} % toggling existential quantifier

\newcommand{\tga}{\mbox{\hspace{1pt}\Large $\wedge$\hspace{-1.84mm}\raisebox{0.02mm}{\rule{0.13mm}{3.0mm}}\hspace{6pt}}} % toggling universal quantifier

\newcommand{\tgpst}{\mbox{\raisebox{-0.01cm}{\scriptsize $\wedge$}\hspace{-4pt}\raisebox{0.06cm}{\small $\mid$}\hspace{2pt}}} % toggling recurrence

\newcommand{\tgpcost}{\mbox{\raisebox{0.12cm}{\scriptsize $\vee$}\hspace{-3.8pt}\raisebox{0.04cm}{\small $\mid$}\hspace{2pt}}} % toggling corecurrence

\newcommand{\tgst}{\mbox{\raisebox{-0.05cm}{$\circ$}\hspace{-0.12cm}\raisebox{0.05cm}{\small $\mid$}\hspace{2pt}}} % toggling-branching recurrence

\newcommand{\tgcost}{\mbox{\raisebox{0.12cm}{$\circ$}\hspace{-0.12cm}\raisebox{0.04cm}{\small $\mid$}\hspace{2pt}}} % toggling-branching corecurrence

\newcommand{\tgpi}{\mbox{\hspace{2pt}\raisebox{0.033cm}{\tiny $>$}\hspace{-0.28cm} \raisebox{-2.3pt}{\LARGE --}\hspace{2pt}}} % toggling-recurrence-based implication

\newcommand{\tgbi}{\mbox{\hspace{2pt}$\circ$\hspace{-0.26cm} \raisebox{-2.3pt}{\LARGE --}\hspace{2pt}}} % toggling-branching-recurrence-based implication

\newcommand{\etc}{{\em etc}}
\newcommand{\cf}{{\em c.f.}}

\begin{abstract}
 Knowledge Representation (KR) is traditionally based on
 the logic of facts, expressed in boolean logic. However,
facts about an agent can also be seen as a set of {\it accomplished}
tasks by the agent.

This paper proposes a new approach to KR: 
the notion of {\it task logical} KR based on Computability Logic.  
This notion allows the user to represent both accomplished tasks and
{\it accomplishable} tasks by the agent.
This notion allows us to build sophisticated KRs about many interesting agents,  
which have  not been supported by previous logical languages.

\end{abstract}

%% </local definitions here>

\section{Introduction}\label{sec:intro}

Traditional acquaintance with knowledge representation (KR) relates to the boolean logic including
classical logic, modal
logic and linear logic \cite{girard87tcs,hodas92ic}.
 Within this setting,  knowledges are expressed as a
logic of facts.  Many KRs in AI textbooks and papers \cite{Kwon08} 
have been written
in boolean logic.
However, boolean logic is too simple to  represent an important aspect of
knowledge, \ie, 
 tasks that can be accomplished by the agent.
In particular, boolean logic is  awkward to use in 
representing accomplishable tasks by many interesting agents.

 It is possible to expand knowledge about an agent  by employing a task/game logic called computability logic (CL) \cite{Jap03, Jap08}, a  powerful logic which is built
around the notion of success/failure.
 CL is a logic of task in which accomplishable tasks can be easily represented.
Consequently, CL can express both deterministic (true/false) and nondeterministic task 
(success/failure) in a
concise way.  
 The task logic offers many new, essential logical operators including
parallel conjunction/disjunction, sequential conjunction/disjunction, choice conjunction/disjunction,
\etc.

This paper proposes to use CL as an KR language. The distinguishing feature
of CL is that now knowledge about an agent include  new, sophisticated tasks that have not
been supported by previous logical languages. While CL is an excellent KR language, it is
based on the first-order logic.  We also consider its higher-order extension where 
first-order
terms are replaced by higher-order terms. It is well-known that higher-order terms can describe
objects of function types including programs and formulas. Higher-order terms have proven
useful in many metalanguage applications such as theorem proving.

The remainder of this paper is structured as follows. We discuss a new way of defining
algorithms in the next section. In Section \ref{sec:modules}, we
present some examples.
Section~\ref{sec:conc} concludes the paper.

\section{Task Logical  KR}\label{sec:logic}

 A {\it task logical} knowledge representation and reasoning (KRR)  is of the form

 \[ \sequent{c:T}{T_1} \]
 
\noindent where $c:T$ represents  an agent $c$ who can do task $T$ and $T_1$ is a query.
In the traditional developments of KR,  $T$ is  limited
to facts or  accomplished tasks.
Accomplishable tasks  are totally ignored. 
In KR, however,  representing accomplishable tasks is  desirable quite often. 
 Such examples include many interesting agents including coffee vending machine,
  many OS processes, lottery tickets, \etc.

To define the class of accomplishable tasks, we need a specification language. 
 An ideal  language would support an optimal
 translation of the tasks.
  We argue that a reasonable, high-level translation of the
tasks can be achieved via computability logic(CL)\cite{jap02,Jap03}. 
An advantage of CL over
other formalisms such as 
sequential pseudocode, linear logic\cite{girard87tcs}, \etc,  is that 
it can optimally encode a number of essential  tasks: nondeterminism, updates, \etc. Hence the main advantage of CL over 
other formalisms is the minimum (linear) size of the encoding.
 
We consider here a higher-order version of CL.
The logical language  we consider in this paper is built based on
   a simply-typed lambda calculus.  Although types
 are strictly necessary, we will
omit these here because  their identity is not relevant in this paper.
An atomic formula is $(p\ t_1\ldots t_n)$
where $p$ is a (predicate) variable or non-logical constant and each $t_i$ is a lambda term. 

The basic operator in CL
is the reduction of the form $c: A \ra B$. This expression
means that the task  $B$ can be reduced to another task $B$.
The expression $c: A\land B$ means that the agent $c$ can perform  two tasks $A$ and $B$ in 
parallel.
The expression  $!A$ means that the agent  can perform the task $A$ repeatedly. 
The expression $c: A \adc B$ means  that the agent $c$ can perform  either task $A$ or $B$, 
regardless of what
the machine chooses.
The expression $c: \ada x A(x)$ means  that the agent $c$ can perform the task $A$, regardless of what
the machine chooses for $x$. The expression $c: A \add B$ means  that the agent $c$ can choose
and perform a true disjunct between $A$ and $B$.

The expression $c: \ade x A(x)$ means that the agent can choose a 
right value for $x$ so that it can perform the task $A$.
 We point the reader to \cite{Jap03,Jap08} to find out more about the
whole calculus of CL.

\section{Examples }\label{sec:modules}
% we need full linear logic + diffentia + proba
The notion of CL makes KR versatile
compared to traditional approach. 
As an  example, we present an agent $c$ who can compute the factorial
function.  This task can be
defined as follows in English:

\begin{numberedlist}

\item   $c$ can either claim that $fac(0,1)$ holds, or

\item  can replace $fac(X,Y)$  by $fac(X+1,XY+Y)$.

\end{numberedlist}
\noindent
It is shown below that the above description can be translated into CL
formulas.
The following is a CL translation of the above knowledge,
where  the reusable action is preceded with $!$.
%We assume that $fact$ is a constant of type $int \ra int \ra o$.
Note that our version use $\adc$ which dynamically creates/destroys facts.
\begin{exmple}
$c: !\ (fac\ 0\ 1)\ \adc\   \ada x\ada y\ ((fac\ x\ y)\ \ra\ (fac\ x+1\ xy+y))$.\\
\end{exmple}
\noindent A task of answering queries 
is typically given  in the form of a query relative to
agents. Computation tries to solve the query with respect to the agent $c$. 
As an example, executing $\sequent{agent\ c}{\ada y \ade z fac(y,z)}$ would involve the
user choosing a value, say 5,  for $y$. This eventually results
in the initial resource $fac(0,1)$ being created and then transformed to $fac(1,1)$, then 
to $fac(2,2)$, and so on. It  will finally produce the desired
result $z = 120$  using the second conjucnt five times.

An example of interactive, accomplishable tasks is provided by the 
following  agent $t$ which is a lottery ticket. 
The ticket is represented as $ 0 \add\  1M$ which indicates that it
has two possible values, nothing or one million dollars.

 The following is a CL translation of the above agent.

\begin{exmple}
\>$t:   0 \add\  1M$.\\
\end{exmple}
\noindent  Now we want to   obtain a final value of $t$. 
This  task is  represented by the query $t$.
 Now executing the program   
$\sequent{agent\ t}{agent\ t}$  would  produce the following question asked by the agent in the task of $ 0 \add\  1M$
 in the program: ``how much  is the final value?''. 
The user's response would be zero dollars. 
This move brings the task down to $\sequent{0}{agent\ t}$. Executing 
$\sequent{0}{agent\ t}$ would  require the 
machine to choose zero dollars in $ 0 \add\  1M$ for a success.

An example of parallel tasks is provided by the agent $b$ which
consists of  two (sub)agents $c$ and $d$ working at a fastfood restaurant.  
The agent $c$ waits for a customer to pay money(at least three dollars), and
then generates a hamburger set consisting of a hamburger,
      a coke and a change. 
The agent $d$ waits for a customer to pay money(at least four dollars), and
then generates a fishburger set consisting of a fishburger,
      a coke and a change. 

The following is a CL translation of the above algorithm.

\begin{exmple}
\>$c: ! \ada x (\geq (x,3)\ \ra\ m(ham) \land  m(coke)\land m(x-3)) \land$ \\
\>$d: ! \ada x (\geq (x,4)\ \ra m(fi) \land  m(coke)\land m(x-4))$.\\
\end{exmple}
\noindent  Now we want to execute $c$ and $d$ in parallel to   obtain a hamburger set and then
a fishburger set by interactively paying money to $c$ and $d$. 
This interactive task is  represented by the query $c\land d$.
 Now executing the program   
$\sequent{agent\ c, agent\ d}{agent\ c \land agent\ d}$  would  produce the following question asked by the agent in the task of 
$c$: ``how much do you want to pay me?''. The user's response would be five dollars. This move brings the task down to $m(ham) \land  m(coke)\land m(\$2)$ which would be a success. The task of $d$ would proceed similarly.

\newcommand{\prov}{pv}
\renewcommand{\ext}{pv_i}

As an example of higher-order KR, consider the interpreter for Horn clauses.
 It is described
by $G$- and $D$-formulas given by the syntax rules below:
\begin{exmple}
\>$G ::=$ \>   $A \sep  G\  and\  G \sep   some\  x\ G $ \\   \\
\>$D ::=$ \>  $A  \sep G\ imp\ A\ \sep all\ x\ D \sep  D\ and\ D$\\
\end{exmple}
\noindent
In the rules above,   
$A$  represents an atomic formula.
A $D$-formula  is called a  Horn
 clause. The expression $some\ x\ G$ involves bindings. We represent such
objects using lambda terms. For example, $all\ x\ p(x)$ is represented as
$all\ \lambda x (p\ x)$.
 
In the algorithm  to be considered, $G$-formulas will function as 
queries and $D$-formulas will constitute  a program. 

 We will  present an operational 
semantics for this language based on \cite{hol}. 
 Note that execution  alternates between 
two phases: the goal-reduction phase 
and the backchaining phase. Following Prolog's syntax, we assume that
names beginning with uppercase letters are quantified by $\ada$.

\begin{defn}\label{def:semantics}
Let $G$ be a goal and let $D$ be a program.
Then the notion of   executing $\lb D,G\rb$ -- $\prov\ D\ G$ -- 
 is defined as follows:
\begin{numberedlist}

\item  $bc\ D\ A\ A\ \adc$ \% This is a success.

\item   $\prov\ D\ G_1\ \ra\ bc\ D\ (G_1\ imp\ A)\ A)\ \adc $

\item   $bc\ D\ (D\ X)\ A\  \ra\ bc\ D\ (all\ D)\ A\ \adc $

\item  $bc\ D\ D_1\ A  \lor
  bc\ D\ D_2\ A\ \ra\  bc\ D\ (D_1\ and\ D_2)\ A\ \adc $

\item   $atom\ A \land\ bc\ D\ D\  A\ \ra\ \prov\ D\ A\ \adc $ \%  change to backchaining phase.
\item  $\prov\ D\ G_1  \land
  \prov\ D\ G_2\ \ra\ \prov\ D\ (G_1\ and\ G_2)\ \adc $

\item  $\prov\ D\ (G\ X)\ \ra\ \prov\ D\ (some\  G)$.

% This goal behaves as exclusive-OR. 

\end{numberedlist}
\end{defn}

\noindent  
In the rules (3) and (7), the symbol $X$  will be instantiated by a term.
In this context, consider the query $\prov\ (p\ a)\ (some\ \lambda x (p\ x))$.
In solving this query, $\prov\ (p\ a)\ (p\ a)$ will be formed and eventually 
solved.

The examples presented here have been of a simple nature. They are, however,
sufficient for appreciating the attractiveness of the algorithm development
process proposed here. We point the reader to \cite{MN87slp,lp2.7,hol} for
 more examples.

\section{Conclusion}\label{sec:conc}

 Knowledge representation  is traditionally based on
 the logic of facts, expressed in boolean logic.
This paper proposed a  new, task logical approach to KR. 
This approach allows us to build sophisticated KRs about many interesting agents,  
which have not been supported by previous boolean logical languages.

Our ultimate interest is in a procedure for carrying out computations
of the kind described above.  Hence it is important to realize this 
CL interpreter in an efficient way, taking advantages of  some techniques  discussed in 
\cite{ban02,CHP96,hodas92ic}. 

\section{Acknowledgements}

This paper was supported by Dong-A University Research Fund.

\bibliographystyle{ieicetr}% bib style

\begin{thebibliography}{1}

\bibitem{ban02}
M. Banbara.
\newblock {\em Design and implementation of linear logic programming languages}.
\newblock  Ph.D. Dissertation, Kobe University, 2002.

\bibitem{CHP96}
Iliano Cervesato, Joshua~S. Hodas, and Frank Pfenning.
\newblock Efficient resource management for linear logic proof search.
\newblock In {\em Proceedings of the 1996 Workshop on Extensions of Logic
  Programming}, LNAI 1050, pages 67 -- 81.


\bibitem{girard87tcs}
Jean-Yves Girard.
\newblock Linear logic.
\newblock {\em Theoretical Computer Science}, 50:1--102, 1987.

\bibitem{hodas92ic}
Joshus Hodas and Dale Miller.
\newblock Logic programming in a fragment of intuitionistic linear logic.
\newblock {\em Journal of Information and Computation}, 1994.
\newblock Invited to a special issue of submission to the 1991 LICS conference.

\bibitem{jap02}
G. Japaridze.
\newblock The logic of tasks.
\newblock {\em Annals of Pure and Applied Logic}, 117:263--295, 2002.

\bibitem{Jap03}
G. Japaridze.
\newblock Introduction to computability logic.
\newblock {\em Annals of Pure and Applied Logic}, 123:1--99, 2003.


\bibitem{Jap08}
G.~Japaridze.  
\newblock Sequential operators in computability logic.
\newblock {\em Information and Computation}, vol.206, No.12, pp.1443-1475, 2008.  
\bibitem{Kwon08}
K.~Kwon and D. Kang.  
\newblock Extending Logicweb via Hereditary Harrop Formulas.
\newblock {\em IEICE Transactions on Information and Systems}, vol.E91-D, No.6,
 pp.1827-1829, 2008.  

\bibitem{MN87slp}
D. Miller and G. Nadathur. 1987.
\newblock A logic programming approach to manipulating formulas and programs.
\newblock In {\em IEEE Symposium on Logic Programming}, {S.~Haridi}, Ed. IEEE
  Computer Society Press, 379--388.

\bibitem{lp2.7}
D. Miller and G. Nadathur. 1988.
\newblock $\lambda${P}rolog version 2.7.
\newblock Distributed in C-Prolog and Quintus Prolog source code.

\bibitem{hol}
D. Miller and G. Nadathur. 2012.
\newblock Programming with higher-order logic.
\newblock Cambridge University Press.


\end{thebibliography}

\end{document}